\documentclass{article} % For LaTeX2e
\usepackage{iclr2024_conference,times}

\usepackage{amsmath,amsfonts,bm}

\def\eqref#1{equation~\ref{#1}}
\def\1{\bm{1}}

\DeclareMathAlphabet{\mathsfit}{\encodingdefault}{\sfdefault}{m}{sl}
\SetMathAlphabet{\mathsfit}{bold}{\encodingdefault}{\sfdefault}{bx}{n}

\usepackage{hyperref}
\usepackage{url}

\usepackage[T1]{fontenc}

\usepackage{arydshln}
\usepackage{booktabs}
\usepackage{enumitem}
\usepackage{balance}
\usepackage{combelow}
\usepackage{tabularx}
\usepackage{cite}
\usepackage{algpseudocode}
\usepackage{amsmath}
\definecolor{azure}{rgb}{0.0, 0.5, 1.0}
\usepackage{wrapfig}
\usepackage[stable]{footmisc}

\usepackage{multirow}
\usepackage{makecell}
\usepackage{mathtools}
\usepackage{algorithm}
\usepackage{graphicx}
\usepackage{algorithmicx}
\usepackage{eqparbox,array}

\usepackage{xcolor,colortbl}

\usepackage{subcaption}
\usepackage{caption}
\usepackage{bbm}
\usepackage{mwe} % qualitative examples
\usepackage{calc} % qualitative examples
\usepackage{pifont} % checkmark in ablation
\newcommand{\method}{\mbox{{CAMA}}\xspace}
\newcommand{\methodfull}{\mbox{{Confidence-Aware Moving Average}}\xspace}
\newcommand{\behaviorIL}{\mbox{{Behavior-IL}}\xspace}
\newcommand{\behaviorILfull}{\mbox{{Behavior Incremental Learning}}\xspace}
\newcommand{\environmentIL}{\mbox{{Environment-IL}}\xspace}
\newcommand{\environmentILfull}{\mbox{{Environment Incremental Learning}}\xspace}

\newcommand{\eg}{\textit{e.g.}\xspace}
\newcommand{\ie}{\textit{i.e.}\xspace}
\newcommand{\etc}{\textit{etc.}\xspace}

\newcommand{\rev}[1]{\textcolor{black}{#1}}

\title{Online Continual Learning for \\Interactive Instruction Following Agents}

\author{Byeonghwi Kim$^{1,}$\thanks{Equal contribution. $^\dagger$Corresponding author. Most of the work is done while JC is with Yonsei University.}~~~~Minhyuk Seo$^{1,*}$~~~~Jonghyun Choi$^{2,\dagger}$ \\
$^{1}$Yonsei University~~~$^{2}$Seoul National University\\
\texttt{\{byeonghwikim,dbd0508\}@yonsei.ac.kr}, \texttt{jonghyunchoi@snu.ac.kr}
}

\iclrfinalcopy % Uncomment for camera-ready version, but NOT for submission.
\begin{document}

\maketitle

\begin{abstract}
    In learning an embodied agent executing daily tasks via language directives, the literature largely assumes that the agent learns all training data at the beginning.
    We argue that such a learning scenario is less realistic since a robotic agent is supposed to learn the world continuously as it explores and perceives it.
    To take a step towards a more realistic embodied agent learning scenario, we propose two continual learning setups for embodied agents; learning new behaviors (Behavior Incremental Learning, Behavior-IL) and new environments (Environment Incremental Learning, Environment-IL) %learning new behaviors (\behaviorILfull, \behaviorIL) and new environments (\environmentILfull, \environmentIL).
    For the tasks, previous \emph{`data prior'} based continual learning methods maintain logits for the past tasks.
    However, the stored information is often insufficiently learned information and requires task boundary information, which might not always be available.
    Here, we propose to update them based on confidence scores without task boundary information \rev{during training} (\ie, \emph{task-free}) in a moving average fashion, named \methodfull (\method).
    In the proposed \behaviorIL and \environmentIL setups, our simple \method outperforms prior \rev{state of the art} in our empirical validations by noticeable margins.
    The project page including codes is {\color{magenta} \url{https://github.com/snumprlab/cl-alfred}}.
\end{abstract}

\section{Introduction}
\label{sec:introduction}

Recent advances in computer vision, natural language processing, and embodied AI have led to various benchmarks for robotic agents, encompassing navigation~\citep{habitat19iccv,deitke2020robothor,anderson2018vision,krantz2020navgraph}, object interaction~\citep{zhu2017visual,misra2017mapping,weihs2021visual,ehsani2021manipulathor}, and interactive reasoning~\citep{embodiedqa,gordon2018iqa}.
To create more realistic agents, challenging benchmarks~\citep{shridhar2020alfred,padmakumar2022teach} require all of these tasks to complete complex tasks based on language directives.

However, most embodied AI literature assumes that all training data are available from the outset but it may be unrealistic as agents may encounter novel behaviors or environments after deployment.
To learn new behaviors and environments, continual learning might be necessary for post-deployment. % beyond their initial training data.

To learn new tasks, one may finetune the agents. 
But the finetuned agents would suffer from catastrophic forgetting that loses previously learned knowledge~\citep{mccloskey1989catastrophic,ratcliff1990connectionist}.
To mitigate such forgetting, \citep{powers2022cora} introduced a continual reinforcement learning framework that incrementally updates agents for new tasks and evaluates their knowledge of current and past tasks.
However, this operates in a simplified task setup of~\citep{shridhar2020alfred}, excluding natural language understanding and object localization. %, which can limit the agents capabilities.

\begin{figure*}[h!]
    \centering
    \includegraphics[width=.97\linewidth]{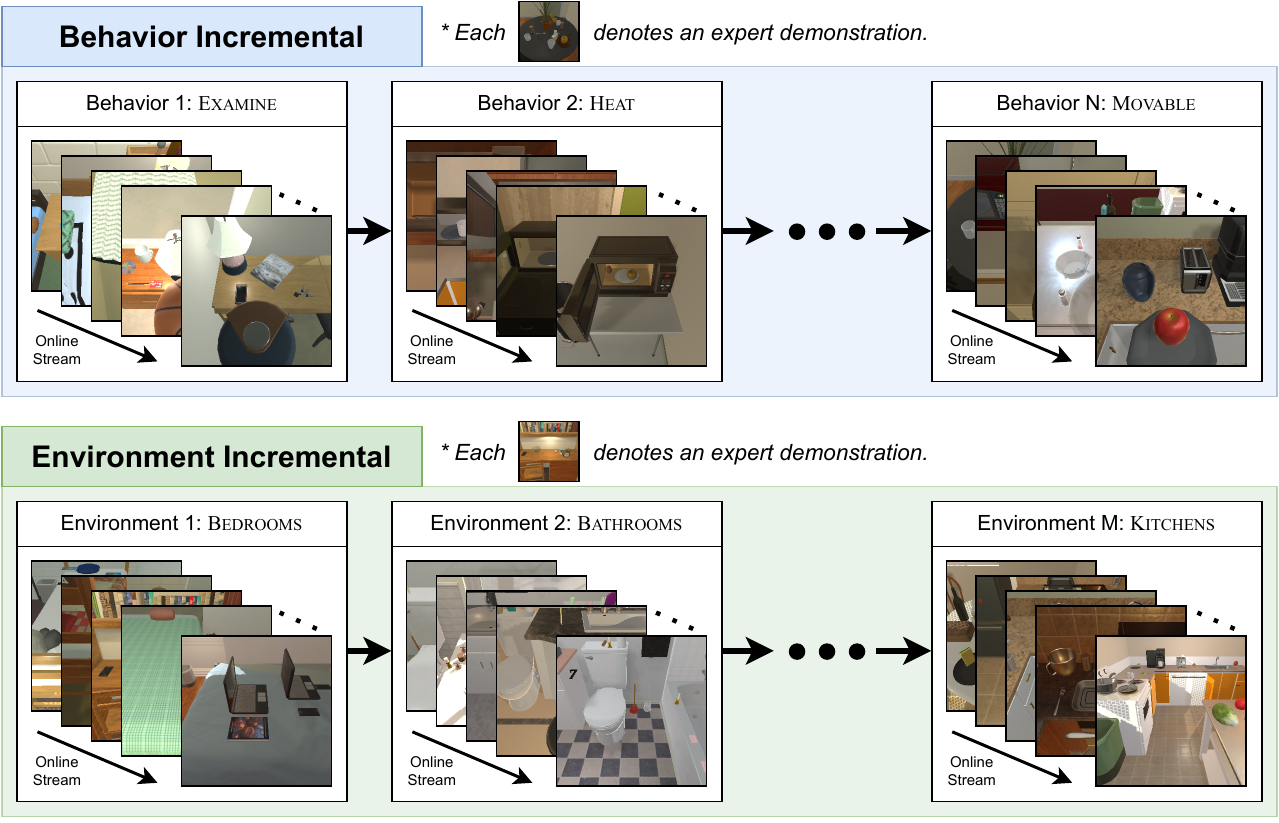}
    \vspace{-.5em}
    \caption{
        \textbf{Proposed two incremental learning setups.}
        In the `Behavior Incremental' setup, the agent is expected to learn new behaviors while preserving previously learned knowledge.
        In the `Environment Incremental' setup, the agent is expected to learn tasks in new environments with the preservation of previously learned knowledge.
        Note that each image in the figure denotes an expert demonstration (\ie, a sequence of frames with natural language instructions followed by a corresponding sequence of actions and object class labels).
    } 
    \label{fig:overview_setup}
    \vspace{-1em}
\end{figure*}

Taking a step forward to bring the instruction following task to real-world scenarios, we propose two continual learning scenarios for embodied agents: Behavior Incremental Learning (\behaviorIL) and Environment Incremental Learning (\environmentIL) as depicted in Figure~\ref{fig:overview_setup}.
In \behaviorIL, the robot learns behaviors incrementally.
For example, it may initially learn object movement and subsequently acquire the skill of object heating.
In \environmentIL, instead of being limited to specific scenes such as bathrooms, the robot progressively learns to perform behaviors in diverse environments such as kitchens and bedrooms.

In the continual learning literature, significant progress~\citep{mai2022online,biesialska2020continual} has been made in addressing continual learning by storing models learned in the previous task of extracting information about past data, requiring a substantial storage cost~\citep{zhou2022model}.
To address this, \citet{buzzega2020dark,boschini2022class} propose to store logits of past models for knowledge distillation, reducing storage costs while maintaining learning efficacy.
However, the stored logits may be the underfitted or insufficiently learned solution as the model has not sufficiently trained in the early stage of learning, hindering the effective use of prior knowledge.
Moreover, such an update often exploits task boundary information that might not always be available, especially in the cases of streamed data without explicit task boundaries~\citep{shanahan2021encoders,kohonline}.

To develop continuously updating embodied agents, we propose to update logits by combining the previously stored logits and the newly obtained ones in the moving average, call \emph{`Confidence-Aware Moving Average'} (\method).
In particular, we dynamically determine the moving average coefficients based on the classification confidence scores inferred by the agents as indicators of the `quality' of the newly obtained logits (\ie, how much they contain accurate knowledge of the corresponding tasks), as empirically observed that high confidence tends to have high accuracy in Figure~\ref{fig:conf_acc}.

\vspace{-.5em}
\paragraph{Contributions.} We summarize our contributions as follows:
\vspace{-1em}
\begin{itemize}[leftmargin=10pt]
\setlength\itemsep{-0.2em}
    \item We propose behavior incremental (\behaviorIL) and environment incremental (\environmentIL) setups for online continual learning for interactive instruction following agents.
    \item We propose \methodfull (\method) that dynamically determines coefficients for logit update to prevent logits from being outdated for effective knowledge distillation.
    \item Our proposed method outperforms comparable methods in most metrics with noticeable margins.
\end{itemize}

\section{Related Work}
\label{sec:related_work}

% \vspace{-.25em} 
\paragraph{Continual learning setups.}
Continual Learning (CL) are typically categorized into two main scenarios: \emph{offline}~\citep{rebuffi2017icarl, ewc, chaudhry2018riemannian, wu2019large} and \emph{online}~\citep{koh2022online, buzzega2020dark, mai2022online, kohonline, Aljundi2019OnlineCL}, based on the frequency with which the model accesses task data.
In the offline setup, data from the current task are used for training multiple times, but this often requires significant memory capacity to store all task data~\citep{hayes2022online}.
On the other hand, online CL involves processing individual or small batches of samples, each of which was used only once for training~\citep{koh2022online, aljundi2019gradient}.
Considering memory constraints and the continuous arrival of limited data points over time in practical scenarios, we focus on the online CL setup.

\vspace{-.75em} 
\paragraph{Task-free continual learning setups.}
\rev{Approaches to continual learning can be categorized into task-free methods~\citep{Aljundi2019OnlineCL, kohonline,  ye2022task, koh2022online} and task-aware methods~\citep{ewc, li2017learning, wu2019large, boschini2022class} based on the use of task boundary information during training~\citep{aljundi2018task}.
Several task-aware methods~\citep{ li2017learning, wu2019large} that exploit task boundary information during training distill the knowledge of past tasks from the previously learned model saved in memory.
However, in real-world scenarios, it is often impractical to know the task identity of streamed input data~\citep{aljundi2018task}.
Therefore, even if tasks are discretely defined (\eg, blurry~\citep{prabhu2020gdumb, bang2021rainbow}, disjoint~\citep{parisi2019continual}), the task-free assumption indicates that no task-specific information or identifier is used during training.}

\vspace{-.75em} 
\paragraph{Knowledge distillation in online continual learning.}
Continual learning has made significant progress, employing methods such as replay-based~\citep{wu2019large, rolnick2019experience, prabhu2020gdumb, bang2021rainbow, koh2022online}, distillation~\citep{li2017learning, buzzega2020dark, kohonline}, and regularization~\citep{zenke2017continual, ewc, lesort2019regularization}.
In particular, distillation-based approaches~\citep{li2017learning,kohonline,boschini2022transfer} have been extensively investigated to leverage prior knowledge, but often require substantial memory and additional computation. 
\rev{Memory requirements make} them unsuitable for settings with limited memory in edge devices~\citep{zhou2022model}.

To address these issues, \citep{buzzega2020dark, michieli2021knowledge, boschini2022class} propose using logits instead of storing previous models, saving memory and inference overhead.
However, the method of storing logits in memory without updating may hinder the current model in distilling outdated past information from stored logits, since the stored logits may represent incomplete learning for past tasks.
To tackle this, a recent approach X-DER~\citep{boschini2022class} updates logits through weighted summation with \rev{logits maintained in memory} and those from the current model, preventing the previously stored logit from becoming outdated, as the model updates.

However, \citet{boschini2022class} requires task boundary information during the training process for logit update, making it unsuitable for setups where data arrive in a continuous stream without task boundaries.
In contrast, our approach updates logits based on the agent's confidence without task boundaries, making it suitable for more general setups where we do not have such information.

\vspace{-.75em} \paragraph{Lifelong learning for robotic agents.}
Going beyond relatively straightforward task setups such as image classification, a substantial amount of literature has emerged to construct a more realistic agent capable of incremental learning of novel tasks~\citep{lesort2020continual} in various aspects including learning strategies (\eg, reinforcement learning~\citep{khetarpal2018environments,wolczyk2021continual,xie2022lifelong}, imitation learning~\citep{mendez2018lifelong,gao2021cril}, \etc) and task formulations (\eg, manipulation~\citep{yang2022evaluations,liu2023libero}, walking~\citep{zhou2022forgetting}, \etc).
Typically, most of this research has focused mainly on relatively fine-grained manipulation tasks, while the navigation aspect~\citep{krantz2020navgraph,deitke2020robothor} has received comparatively less attention.

Concurrently, there is recent literature that delves into the dual aspect of navigation and interaction~\citep{powers2022cora,wang2023voyager} in 3D interactive environments~\citep{ai2thor,fan2022minedojo} to perform more demanding tasks.
In this context, agents are required to become proficient in both navigating and interacting with task-related objects.
Here, the tasks in our proposed continual learning setups are similar to \citep{powers2022cora} that simplifies the task setup of \citep{shridhar2020alfred}.
While \citep{powers2022cora} excludes natural language understanding and object localization, we include them to train agents to complete the desired tasks in the challenging full-fledged interactive instruction following setup, along with navigation and object interaction.

We review more relevant literature and provide extended related work in Sec.~\ref{sec:extended_related_work} for space’s sake.

\section{CL-ALFRED: Continual Learning Setups for Embodied Agents}
\label{sec:benchmark}
Continual learning enables agents to adapt to new behaviors and diverse environments after deployment, mitigating the risk of forgetting previously acquired knowledge.
To foster active research on mitigating catastrophic forgetting, recent literature~\citep{powers2022cora} proposes a benchmark that continuously learns new types of household tasks, but lacks natural language understanding and object localization, potentially limiting the deployability of agents.

To comprehensively address the combined challenges of continuous learning of an agent with natural language understanding and object localization, we use full-fledged interactive instruction following tasks and propose two incremental learning setups, \emph{\behaviorILfull} (\textbf{\behaviorIL}) and \emph{\environmentILfull} (\textbf{\environmentIL}).
We outline our task formulation and detail these incremental learning setups in the following sections.

\subsection{Task Formulation}
\label{subsec:task_definition}
As the ALFRED dataset~\citep{shridhar2020alfred} requires a comprehensive understanding of natural language and visual environments for intelligent agents, we build our continual benchmark on top of it.
The agent is first spawned at a random location in an environment and then given natural language instructions, $l$, that describe how to accomplish a task.
For each time step $t$, the agent takes as input visual observation $v_t$ and predicts an action $y_{a,t}$ and a mask $y_{m,t}$ of an object class $y_{c,t}$ if object interaction is required.
Here, we learn a policy parameterized by $\theta$, $\pi_\theta:x\xrightarrow{}y$, with input $x_t$, \ie, ($v_t$, $l$) and output $y_t$, \ie, ($y_{a,t}$, $y_{m,t}$).
The goal for the policy $\pi_\theta$ is to predict a sequence of actions and object masks to complete the task.
Kindly refer to \citet{shridhar2020alfred} for more details.

Most previous methods~\citep{singh2021factorizing,pashevich2021episodic,min2021film,kim2023context} for object localization utilize a two-stage approach, separating it into object class prediction and mask generation to enhance object localization.
Since mask generation is handled by separate mask generators, however, continual updates of these networks are also required.
Unfortunately, continuously updated instance segmentation models~\citep{joseph2021incremental,cermelli2022modeling} often noticeably underperform jointly trained models.
Here, we only address class prediction, assuming the availability of object masks, leaving the continual updating of mask generators for future work.

\subsection{Continual Learning Setups}
\label{subsec:continual_setup}
There is significant progress in developing agents that can perform the desired tasks through language directives~\citep{krantz2020navgraph,shridhar2020alfred,padmakumar2022teach}.
It is often confronted with new behaviors and environments after being deployed and are required to learn them while maintaining previously learned knowledge.
However, prior methods either presuppose the availability of pre-collected datasets or utilize simplified task configurations~\citep{powers2022cora}.

To address this limitation, we introduce two continual learning setups: 1) \textbf{Behavior Incremental Learning} (\textbf{Behavior-IL}) to incrementally learn \textit{what} to do and 2) \textbf{Environment Incremental Learning} (\textbf{Environment-IL}) to incrementally learn \textit{where} to do, as in Figure~\ref{fig:overview_setup}.
In addition, we focus on online CL, which assumes a more realistic scenario where novel data are encountered in a streaming manner~\citep{Aljundi2019OnlineCL, kohonline, koh2022online} rather than assuming an offline CL in which novel data are provided in chunks of tasks~\citep{wu2019large, saha2021gradient}.
More details about the continual setup can be found in Sec.~\ref{subsec:continual_learning_setups_appendix}.

\subsubsection{\behaviorILfull}
\label{subsubsec:behavior_il}
Behaviors described by instructions may exhibit considerable diversity as novel behaviors may arise over time.
To address this scenario, we propose the \behaviorIL setup that facilitates the agent's incremental learning of new behaviors while retaining the knowledge obtained from previous tasks.

Formally, for a set of behaviors, $\mathcal{T}$, an agent sequentially receives $N_j$ training episodes, $\{s_i^{\tau_j}\}_{i=1}^{N_j}$, for each type of behavior, $\tau_j \in \mathcal{T}$.
When receiving the final episode (\ie, $s_{N_j}^{\tau_j}$) for the current behavior type, $\tau_j$, the agent starts to sequentially receive episodes, $\{s_i^{\tau_{j+1}}\}_{i=1}^{N_{j+1}}$, for the next behavior type, $\tau_{j+1}$.
The episode stream ends with the last training episode, $s_{N_{|\mathcal{T}|}}^{\tau_{|\mathcal{T}|}}$, of the last task type, $\tau_{|\mathcal{T}|}$.

Here, we adopt seven different types of behavior from \citep{shridhar2020alfred}: \textsc{Examine}, \textsc{Pick\&Place}, \textsc{Heat}, \textsc{Cool}, \textsc{Clean}, \textsc{Pick2\&Place}, and \textsc{Movable}.
To ensure the adaptability of agents and avoid favoring particular behavior sequences, we train and evaluate agents using multiple randomly ordered behavior sequences.
Refer to Sec.~\ref{subsec:behavior_il_appendix} for more details about the sequences.

\subsubsection{\environmentILfull}
\label{subsubsec:environment_il}
The \environmentIL setup allows agents to learn the environment incrementally.
In the real world, agents often need to perform actions not only in the environment in which they were initially trained but also in new and different environments that are presented.
For example, the agent may first learn various actions in a kitchen and then subsequently learn the actions in a bathroom.

Formally, for a set of environments, $\mathcal{E}$, an agent sequentially receives $M_k$ training episodes, $\{s_i^{e_k}\}_{i=1}^{M_k}$, for each environment type, $e_k \in \mathcal{E}$.
When receiving the final episode (\ie, $s_{M_k}^{e_k}$) for the current environment type, $e_k$, the agent begins to sequentially receive episodes, $\{s_i^{e_{k+1}}\}_{i=1}^{M_{k+1}}$, for the next environment type, $e_{k+1}$.
This is repeated until it reaches the last environment type, $e_{|\mathcal{E}|}$.

In this setup, we adopt four different types of environments supported by \citep{ai2thor}: \textsc{Kitchens}, \textsc{Livingrooms}, \textsc{Bedrooms}, and \textsc{Bathrooms}.
Like the \behaviorIL setup, we conduct training and evaluation using multiple sequences of randomly ordered environments.
We also provide more details of the multiple environment sequences in Sec.~\ref{subsec:environment_il_appendix}.

However, we observe an imbalance in the training and evaluation episodes between different types of environment~\citep{shridhar2020alfred}, where a majority of them originate from a specific environment type in many instances.
The imbalance can potentially lead to biased learning of a model towards the dominant (\ie, majority) environment type~\citep{chakraborty2020superensemble, zhao2021energy}.
To address the issue, we balance them by simply subsampling the training and evaluation episodes for each environment to match the number of episodes across environment types equally.

\begin{figure*}[t!]
    \centering
    \includegraphics[width=.99\linewidth]{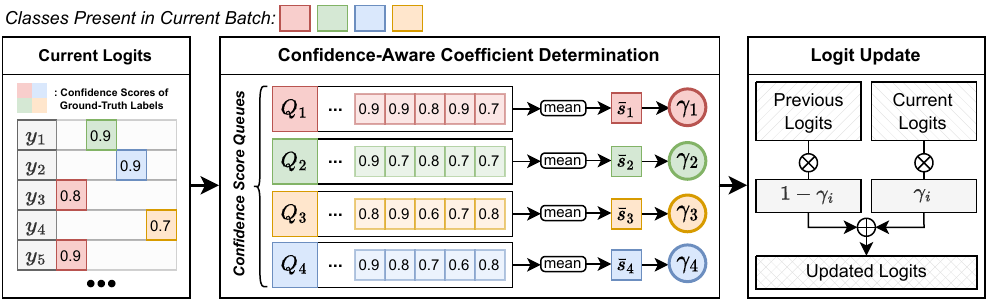}
    \vspace{-.5em}
    \caption{
        \textbf{Proposed \methodfull (\method).}
        `Current Logits' denotes the model's logits obtained from the input samples from the current stream and episodic memory.
        `Previous Logits' denotes logits stored in episodic memory before an update.
        $Q_i$ denotes a queue that stores ground truth confidence scores acquired from the current logits, $y_1, y_2, ...$, for the $i^{th}$ class.
        To obtain $\gamma_i$, we maintain the recent $N$ confidence scores for the $i^{th}$ class and calculate the mean value of the scores followed by a clip function.
        Finally, we use all $\gamma_i$'s to class-wisely weight-sum previously stored logits (\ie, 'Previous Logits') and newly obtained logits from the current stream (\ie, `Current Logits').
    }
    \vspace{-1em}
    \label{fig:overview_approach}
\end{figure*}

\section{Approach}
\label{sec:approach}

\begin{wrapfigure}{r}{0.47\textwidth}
    \vspace{-2em}
    \begin{center}
        \includegraphics[width=\linewidth]{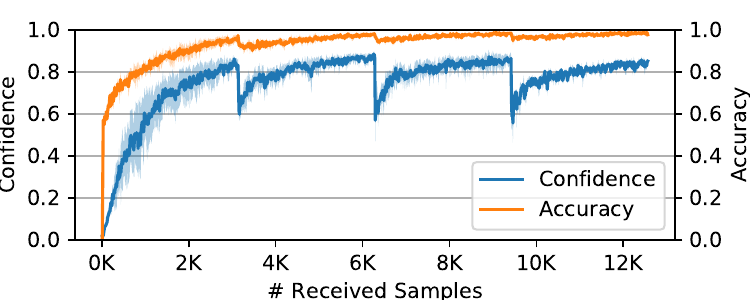}
    \end{center}
    \vspace{-1em}
    \caption{
        \textbf{Confidence and accuracy of logits used for logit update in \method.}
        `Accuracy' denotes the mean of the frame-wise accuracies measured from the newly obtained logits (here, $z'_a$) in Equation~\ref{eq:weighted_sum}.
        `Confidence' denotes the dynamically determined coefficients (here, $\gamma_a$) for the update in Equation~\ref{eq:mean_confidence_score}.
    }
    \vspace{-1em}
    \label{fig:conf_acc}
\end{wrapfigure}

To mitigate catastrophic forgetting, recent approaches~\citep{li2017learning, kohonline, boschini2022transfer} use knowledge distillation with the trained model until past tasks, but this often entails significant memory~\citep{zhou2022model} and computational overhead caused by additional model inference~\citep{prabhu2023computationally}.
Due to the limitations in memory and computation on edge devices~\citep{lee2022carm}, logit distillation methods~\citep{buzzega2020dark, boschini2022class} have been proposed as alternatives to those that store entire models for distillation~\citep{li2017learning, kohonline}, offering improved memory and computation efficiency.
Despite such improved efficiency, some of the logit distillation methods~\citep{buzzega2020dark} often face an outdated logit problem, as the memory-stored logits are not updated to preserve information on previous tasks.

To address this issue, a recent approach~\citep{boschini2022class} attempts to partially update logits stored in the past using the current model.
It uses task boundary information (\eg, input's task identity) during training, but it may not always be available, especially in \emph{task-free} CL setups including our proposed ones.
Towards a general logit-update framework devoid of such information, we update the stored logits using the agent's confidence scores indicating how the newly obtained logits for update contain accurate knowledge of the corresponding tasks, as observed in Figure~\ref{fig:conf_acc}.

\subsection{\methodfull}
\label{subsec:adaptive_moving_average}

As illustrated in Figure~\ref{fig:overview_approach}, the overall process of \methodfull (\method) can be summarized as follows:
Initially, exploiting the model's confidence scores of ground-truth labels, we evaluate the extent to which the model has acquired proficiency in the current samples.
Subsequently, during the computation of the updated logits based on the previous and current logits, we allocate a higher weight to the current logits when exhibiting higher confidence scores, and conversely, assign a higher weight to the previous logits demonstrating lower confidence scores.
For better understanding, we outline the high-level flow of our \method in Algorithm~\ref{algo:sdp} in the appendix.

Following the common practice~\citep{ewc,rolnick2019experience,buzzega2020dark}, we construct an input batch, $[x; x']$, by combining data from both the training data stream $(x, y_a, y_c)\sim\mathcal{D}$ and the episodic memory $(x', y'_a, y'_c, z'_{old,a}, z'_{old,c})\sim\mathcal{M}$, where each $a \in \mathcal{A}$ and $c \in \mathcal{C}$ indicates an action and object class label from the action and object class sets, $\mathcal{A}$ and $\mathcal{C}$, present in the input batch, $[x; x']$.
Here, $x$ represents the input (\ie, images and language directives), $y_a$ and $y_c$ denote the corresponding action and object class labels, and $z'_{old,a}$ and $z'_{old,c}$ refers to the corresponding stored logits.
$z_a$, $z_c$, $z'_a$, and $z'_c$ denote the current model's logits for the input batch.

To prevent the logits maintained in the episodic memory from becoming outdated, we obtain the updated logits, $z'_{new,a}$ and $z'_{new,c}$, by weighted-summing $z'_{old,a}$ and $z'_{old,c}$ with $z'_{a}$ and $z'_{c}$ using coefficient vectors, $\gamma_a$ and $\gamma_c$, using Hadamard product, denoted by $\odot$, as in Equation~\ref{eq:weighted_sum}:
\begin{equation}
    \label{eq:weighted_sum}
    z'_{new,a} = (\mathbf{1} - \gamma_a) \odot z'_{old,a} + \rev{\gamma_a} \odot z'_a, ~~~
    z'_{new,c} = (\mathbf{1} - \gamma_c) \odot z'_{new,c} + \rev{\gamma_c} \odot z'_c.
\end{equation}

To obtain $\gamma_a$ and $\gamma_c$, we first maintain the most recent $N$ confidence scores for each action and object class label for $x$.
Then, to approximate the agent's proficiency in learning tasks over time, we compute the average of the scores associated with each action ($i$) and object class ($j$) label, denoted by $\Bar{s}^a_i$ and $\Bar{s}^c_j$.
We then set each element of $\gamma_a$ and $\gamma_c$, denoted by $\gamma_{a,i}$ and $\gamma_{c,j}$, to $\Bar{s}^a_i$ and $\Bar{s}^c_j$ followed by a CLIP function as in Equation~\ref{eq:mean_confidence_score}:
\begin{equation}
    \label{eq:mean_confidence_score}
    \begin{split}
        \gamma_{a,i} = \alpha_a \text{CLIP} \left( \Bar{s}^a_i - |\mathcal{A}|^{-1}, 0, 1 \right), ~~~
        \gamma_{c,j} = \alpha_c \text{CLIP} \left( \Bar{s}^c_j - |\mathcal{C}|^{-1}, 0, 1 \right),
    \end{split}
\end{equation}
where $\text{CLIP}(x, \min, \max)$ denotes the clip function that limits the value of $x$ from $\min$ to $\max$.
Here, the constants $\alpha_a < 1$ and $\alpha_c < 1$ are introduced to prevent $\gamma_{a,i}$ and $\gamma_{c,j}$ from reaching a value of 1 as this indicates complete replacement of the prior logits with the current logits, which implies that the updated logits would entirely forget the previously learned knowledge.
The inclusion of these constants ensures that some information from the past is retained and not entirely overridden by the current logits during the update process.
In addition, we subtract $|\mathcal{A}|^{-1}$ and $|\mathcal{C}|^{-1}$ enhances the effective utilization of confidence scores in comparison to a `random' selection, which would otherwise be realized by a uniform distribution~\citep{koh2022online}.

\subsection{Model Training}
Given expert demonstrations, $x$ as input, we train our agent, $\pi_\theta$, by minimizing the objective below:
\begin{equation}
    \begin{gathered}
        \min_{\theta} \mathbb{E}_{(x,y)\sim \mathcal{D}}[\mathcal{L}(\pi_\theta(x), y)] + \mathbb{E}_{(x,y)\sim \mathcal{M}}[\mathcal{L}(\pi_\theta(x), y)] + \alpha~\mathbb{E}_{(x,z) \sim \mathcal{M}} [||z - \pi_{\theta}(x)||^2_2],
    \end{gathered}
\end{equation}
where $y$ denotes the ground-truth labels corresponding to $x$ and $z$ the logits maintained in the episodic memory.
We provide more details of the training loss, $\mathcal{L}$, in Sec.~\ref{subsec:model_architecture_and_training} in the appendix. % for space's sake.

\section{Experiments}
\label{sec:experiment}

\paragraph{Evaluation metrics.}
For evaluation of task completion ability, we follow the same evaluation protocol of \citep{shridhar2020alfred}.
The primary metric is the success rate (SR) which measures the ratio of the succeeded episodes among the total ones.
The second metric is the goal-condition success rate (GC) which measures the ratio of the satisfied goal conditions among the total ones.
Furthermore, we evaluated all agents in two splits of environments: \emph{seen} and \emph{unseen} environments which are/are not used to train agents.
We provide more details of the evaluation protocol in Sec.~\ref{subsec:evaluation_metrics}.

To evaluate the last and intermediate performance of continual learning agents, we measure two variations of a metric, $A$: $A_{last}$ and $A_{avg}$.
$A_{last}$ (here, $\text{SR}_{last}$ and $\text{GC}_{last}$) indicates the metric achieved by the agent that finishes its training for the final task.
$A_{avg}$ (here, $\text{SR}_{avg}$ and $\text{GC}_{avg}$) indicates the average of the metrics achieved by the agents that finish their training for each incremental task.

\rev{All the models are trained sequentially over a sequence of behaviors (\behaviorIL) and environments (\environmentIL) and then evaluated over the behaviors and environments that the models have learned so far.
For evaluation, we use episodes different from those used for training.
The same trained models are evaluated in both \emph{seen} and \emph{unseen} environments.
For \emph{seen} and \emph{unseen}, we denote by \emph{seen} the evaluation with the episodes generated from scenes used in training, while we denote by \emph{unseen} the evaluation with the episodes generated from scenes not used in training.}

\vspace{-.5em} \paragraph{Baselines.}
We compare our \method with competitive prior arts in continual learning literature: CLIB~\citep{koh2022online}, DER++~\citep{buzzega2020dark}, ER~\citep{rolnick2019experience}, MIR~\citep{Aljundi2019OnlineCL}, EWC~\citep{ewc}, and X-DER~\citep{boschini2022class}.
In addition, we also compare our \method with two models: `Joint' and `Finetuning'.
`Joint' denotes that the agent is trained with all task data jointly, which works as an upper bound.
`Finetuning' denotes that the agent is fine-tuned for the new tasks or scene types, which can serve as one of the trivial solutions for continual setups.
We provide further explanation for each baseline in Sec.~\ref{subsec:baselines}.
We further detail the model architecture and training used for the methods above in Sec.~\ref{subsec:model_architecture_and_training} for space's sake.

\vspace{-.5em} \paragraph{Implementation details.}
It is a common practice in continual learning literature~\citep{bang2021rainbow, koh2022online, kohonline} to set the size of episodic memory to less than 5\%.
To align with previous works in continual learning, we set the size of the episodic memory to $M=500$ for expert demonstrations, which is approximately $2.38\%$ of the training episodes in the ALFRED benchmark~\citep{shridhar2020alfred}.
For our \method, we empirically set $\alpha_a = 0.99$ and $\alpha_c = 0.99$.
We provide more implementation details such as hyperparameters in Sec.~\ref{subsec:implementation_details} for space's sake.

\subsection{Comparison with State of the Art}
\label{subsec:comparison_with_state_of_the_art}
We present the quantitative results of our \method in Table~\ref{tab:sota_valid_seen}-\ref{tab:sota_valid_unseen}.
As mentioned in Sec.~\ref{subsec:continual_setup}, we train and evaluate our \method and baselines for three random seeds and report the results with their average and standard deviation to avoid favoring particular behavior and environment sequences.
We provide quantitative analyses in various aspects as follows.

\vspace{-.75em} \paragraph{Joint training vs. Finetuning.}
Before investigating the effectiveness of our \method, we first investigate how challenging the proposed \behaviorIL and \environmentIL setups are.
We observe significant performance drops in `Finetuning' compared to `Joint' with 51.0\% and 47.5\% relative drops.
This implies that simply finetuning agents to novel behaviors and environments cannot effectively address the forgetting caused by distribution shifts between behaviors and environments.

\vspace{-.75em} \paragraph{Ours vs. Regularization-based model.}
We observe that our \method achieves better performance than the regularization-based approach (\ie, EWC++) with noticeable margins in both seen and unseen environments for all metrics and setups, indicating that regularizing changes in importance parameters may not effectively prevent forgetting than distilling knowledge from updated logits.

\vspace{-.75em} \paragraph{Ours vs. Rehearsal-based models.}
We observe that our \method outperforms all rehearsal-based approaches (\ie, ER, MIR, and CLIB) for all metrics in both seen and unseen environments in both \behaviorIL and \environmentIL setups.
We believe that this implies that solely depending on sample replay amidst rapid data distribution shifts can result in insufficient task forgetting mitigation and hinder the agent's ability to adapt to novel tasks, ultimately impeding effective task completion.

\vspace{-.75em} \paragraph{Ours vs. Distillation-based models.}
We compare our \method with the distillation-based approaches (\ie, DER and X-DER) to investigate the effectiveness of our logit-update approach.
First, we observe noticeable performance drops in DER, which does not update logits, compared to our \method for all metrics in seen and unseen environments in both \behaviorIL and \environmentIL setups, which highlights the importance of updating logits to prevent them from being outdated.

In addition, we observe that our \method outperforms X-DER, which partially updates logits only for novel classes, with noticeable margins for all metrics and environments in both the \behaviorIL and \environmentIL setups, highlighting the efficacy of our \method.
We note that while X-DER updates logits based on task boundary information during training, our \method does not assume the availability of such information (\ie, task-free), which highlights the generality of our \method.

\newcommand{\mcc}[2]{\multicolumn{#1}{c}{#2}}
\newcommand{\mcp}[2]{\multicolumn{#1}{c@{\hspace{30pt}}}{#2}}
\definecolor{Gray}{gray}{0.90}
\newcolumntype{a}{>{\columncolor{Gray}}r}
\newcolumntype{b}{>{\columncolor{Gray}}c}
\newcommand{\B}[1]{\textcolor{blue}{\textbf{#1}}}

\begin{table*}[t!]
    \centering
    \resizebox{\textwidth}{!}{
        \begin{tabular}{@{}laaaarrrr@{}}
            \toprule
            
            \multirow{4}{*}{Model}
                 & \mcc{8}{\textbf{Validation Seen}} \\
                 \cmidrule{2-9}
                 & \multicolumn{4}{b}{\textbf{\behaviorIL}} & \mcc{4}{\textbf{\environmentIL}} \\
                 & \multicolumn{1}{b}{$\text{SR}_{last}\uparrow$} & \multicolumn{1}{b}{$\text{GC}_{last}\uparrow$}
                 & \multicolumn{1}{b}{$\text{SR}_{avg}\uparrow$} & \multicolumn{1}{b}{$\text{GC}_{avg}\uparrow$}
                 & \multicolumn{1}{c}{$\text{SR}_{last}\uparrow$} & \multicolumn{1}{c}{$\text{GC}_{last}\uparrow$}
                 & \multicolumn{1}{c}{$\text{SR}_{avg}\uparrow$} & \multicolumn{1}{c}{$\text{GC}_{avg}\uparrow$} \\
                             
            \cmidrule{1-9}
            
            Finetuning
                & $9.51 \pm 1.09$ & $20.39 \pm 0.61$
                & $17.07 \pm 0.86$ & $26.11 \pm 0.95$
                & $8.72 \pm 2.12$ & $15.56 \pm 1.29$
                & $16.25 \pm 3.95$ & $21.40 \pm 4.65$
                \\
            \cmidrule{1-9}
            EWC++ %\citep{kirkpatrick2017overcoming}
                & $20.37 \pm 5.19$ & $29.32 \pm 5.92$
                & $22.21 \pm 4.34$ & $30.97 \pm 4.26$
                & $26.79 \pm 2.24$ & $36.79 \pm 1.83$
                & $31.01 \pm 2.76$ & $40.56 \pm 2.22$
                \\             
            ER %\citep{rolnick2019experience}
                & $26.71 \pm 1.49$ & $36.59 \pm 1.36$
                & $27.67 \pm 2.08$ & $36.20 \pm 1.96$
                & $30.28 \pm 1.07$ & $39.15 \pm 0.83$
                & $34.72 \pm 1.56$ & $44.00 \pm 1.52$
                \\
            MIR %\citep{Aljundi2019OnlineCL}
                & $30.27 \pm 1.33$ & $40.14 \pm 2.00$
                & $28.12 \pm 1.78$ & $36.76 \pm 1.73$
                & $27.50 \pm 1.48$ & $36.31 \pm 1.43$
                & $31.81 \pm 0.81$ & $40.94 \pm 0.95$
                \\
            CLIB %\citep{koh2022online}
                & $23.85 \pm 2.02$ & $34.25 \pm 1.81$
                & $23.94 \pm 2.36$ & $32.65 \pm 2.22$
                & $25.47 \pm 1.42$ & $34.63 \pm 1.55$
                & $32.51 \pm 2.40$ & $41.25 \pm 2.34$
                \\
            DER++ %\citep{buzzega2020dark} % all logits
                & $29.15 \pm 1.29$ & $39.39 \pm 1.16$
                & $27.49 \pm 2.27$ & $36.10 \pm 1.92$
                & $28.25 \pm 1.18$ & $36.18 \pm 0.60$
                & $28.68 \pm 2.54$ & $38.01 \pm 3.16$
                \\
            X-DER
                & $28.76 \pm 1.25$ & $38.45 \pm 1.18$
                & $27.21 \pm 2.53$ & $35.88 \pm 2.22$
                & $28.30 \pm 0.87$ & $37.02 \pm 0.61$
                & $29.32 \pm 2.36$ & $39.13 \pm 2.59$
                \\  
            \cmidrule{1-9}
            \method w/o D.C.
                & $30.54 \pm 1.27$ & $39.92 \pm 1.50$
                & $\textbf{29.89} \pm \textbf{2.32}$ & $38.16 \pm 2.71$
                & $\textbf{31.55} \pm \textbf{0.87}$ & $\textbf{39.29} \pm \textbf{1.03}$
                & $34.49 \pm 2.29$ & $43.28 \pm 2.17$
                \\

            \textbf{\method (Ours)}
                & $\textbf{30.71} \pm \textbf{0.78}$ & $\textbf{40.85} \pm \textbf{0.73}$
                & $29.67 \pm 2.66$ & $\textbf{38.17} \pm \textbf{2.34}$
                & $29.48 \pm 0.27$ & $38.13 \pm 0.85$
                & $\textbf{35.09} \pm \textbf{1.92}$ & $\textbf{44.02} \pm \textbf{2.21}$    
                \\
            
            \cmidrule{1-9}
            
            Joint
                & $60.47 \pm 0.33$ & $65.77 \pm 0.78$
                & $-$ & $-$
                & $56.25 \pm 0.89$ & $62.13 \pm 0.84$
                & $-$ & $-$
                \\
            
            \bottomrule
        \end{tabular}
    }
    \vspace{-.5em}
    \caption{
        \textbf{Comparison with state-of-the-art methods (validation seen).}
        The highest value per metric is in \textbf{bold}.
        We report the averages and standard deviations of multiple runs for random seeds as in Sec.~\ref{subsec:continual_setup}.
    }
    \vspace{-.5em}
    \label{tab:sota_valid_seen}
\end{table*}

\begin{table*}[t!]
    \centering
    \resizebox{\textwidth}{!}{
        \begin{tabular}{@{}laaaarrrr@{}}
            \toprule
            
            \multirow{4}{*}{Model}
                 & \mcc{8}{\textbf{Validation Unseen}} \\
                 \cmidrule{2-9}
                 & \multicolumn{4}{b}{\textbf{\behaviorIL}} & \mcc{4}{\textbf{\environmentIL}} \\
                 & \multicolumn{1}{b}{$\text{SR}_{last}\uparrow$} & \multicolumn{1}{b}{$\text{GC}_{last}\uparrow$}
                 & \multicolumn{1}{b}{$\text{SR}_{avg}\uparrow$} & \multicolumn{1}{b}{$\text{GC}_{avg}\uparrow$}
                 & \multicolumn{1}{c}{$\text{SR}_{last}\uparrow$} & \multicolumn{1}{c}{$\text{GC}_{last}\uparrow$}
                 & \multicolumn{1}{c}{$\text{SR}_{avg}\uparrow$} & \multicolumn{1}{c}{$\text{GC}_{avg}\uparrow$} \\
                             
            \cmidrule{1-9}
            
            Finetuning
                & $1.18 \pm 1.09$ & $12.09 \pm 1.65$
                & $3.03 \pm 1.29$ & $13.95 \pm 1.01$
                & $2.01 \pm 0.86$ & $11.20 \pm 1.92$
                & $2.90 \pm 2.16$ & $13.53 \pm 4.33$
                \\
                
            \cmidrule{1-9}
            EWC++ %\citep{kirkpatrick2017overcoming}
                & $8.50 \pm 2.15$ & $21.63 \pm 3.60$
                & $8.33 \pm 1.33$ & $20.71 \pm 1.99$
                & $11.61 \pm 1.29$ & $28.47 \pm 0.83$
                & $12.37 \pm 1.30$ & $29.90 \pm 1.30$
                \\
            ER %\citep{rolnick2019experience}
                & $9.43 \pm 1.25$ & $24.22 \pm 1.54$
                & $9.47 \pm 1.51$ & $22.79 \pm 1.39$
                & $11.44 \pm 1.36$ & $29.11 \pm 0.96$
                & $14.25 \pm 1.47$ & $31.98 \pm 1.64$
                \\
            MIR %\citep{Aljundi2019OnlineCL}
                & $11.01 \pm 1.16$ & $25.31 \pm 1.14$
                & $10.88 \pm 1.51$ & $24.20 \pm 1.45$ 
                & $12.01 \pm 0.61$ & $29.67 \pm 0.47$
                & $12.58 \pm 0.80$ & $29.11 \pm 1.26$
                \\
            CLIB %\citep{koh2022online}
                & $8.26 \pm 1.03$ & $22.00 \pm 1.31$
                & $8.56 \pm 0.66$ & $21.03 \pm 1.15$
                & $10.46 \pm 1.18$ & $27.40 \pm 0.78$
                & $11.95 \pm 1.51$ & $29.93 \pm 2.05$
                \\
            DER++ %\citep{buzzega2020dark} % all logits
                & $13.16 \pm 5.56$ & $28.70 \pm 5.63$
                & $10.60 \pm 4.04$ & $24.94 \pm 2.69$
                & $10.29 \pm 1.05$ & $26.90 \pm 1.31$
                & $10.25 \pm 1.72$ & $26.83 \pm 1.97$
                \\
            X-DER
                & $12.59 \pm 1.92$ & $28.10 \pm 2.05$
                & $12.04 \pm 1.56$ & $25.50 \pm 1.48$
                & $10.75 \pm 1.15$ & $28.37 \pm 1.05$
                & $11.14 \pm 1.38$ & $28.56 \pm 1.44$
                \\
            \cmidrule{1-9}
            \method w/o D.C.
                & $\textbf{14.06} \pm \textbf{1.20}$ & $28.33 \pm 1.58$
                & $12.52 \pm 1.46$ & $26.02 \pm 1.38$
                & $13.57 \pm 1.25$ & $29.54 \pm 1.41$
                & $12.78 \pm 0.57$ & $29.76 \pm 0.84$
                \\
                
            \textbf{\method (Ours)}
                & $13.64 \pm 0.94$ & $\textbf{28.75} \pm \textbf{0.92}$
                & $\textbf{14.19} \pm \textbf{1.38}$ & $\textbf{27.30} \pm \textbf{1.38}$
                & $\textbf{14.60} \pm \textbf{0.43}$ & $\textbf{30.99} \pm \textbf{0.75}$
                & $\textbf{15.67} \pm \textbf{0.77}$ & $\textbf{33.40} \pm \textbf{1.45}$
                \\
            
            \cmidrule{1-9}
            
            Joint
                & $24.60 \pm 0.96$ & $38.24 \pm 1.55$
                & $-$ & $-$
                & $19.73 \pm 2.31$ & $39.02 \pm 0.53$
                & $-$ & $-$
                \\
            
            \bottomrule
        \end{tabular}
    }
    \vspace{-.5em}
    \caption{
        \textbf{Comparison with state-of-the-art methods (validation unseen).}
        The highest value per metric is in \textbf{bold}.
        We report averages and standard deviations of multiple runs for random seeds as in Sec.~\ref{subsec:continual_setup}.
    }
    % \vspace{-.5em}
    \label{tab:sota_valid_unseen}
\end{table*}

\begin{figure*}[t!]
    \centering
    \includegraphics[width=\linewidth]{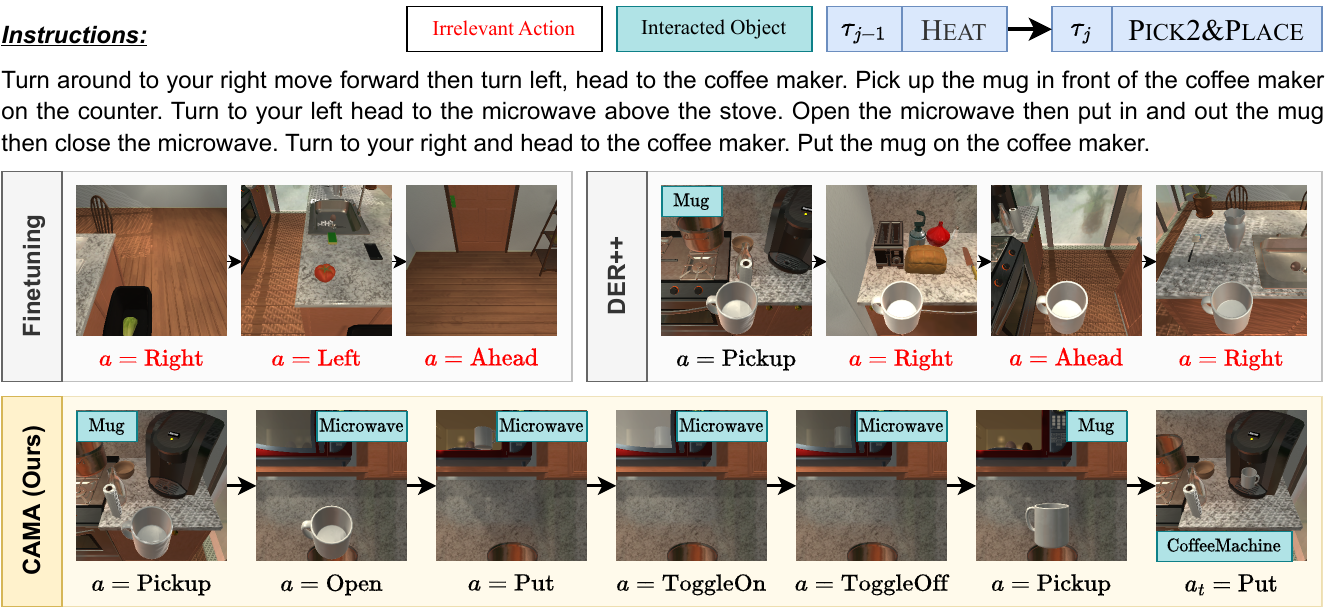}
    \caption{
        \textbf{Qualitative analysis of the proposed method (\behaviorIL).}
        The agent, having already acquired knowledge of the behavior $\tau_{j-1} = \textsc{Heat}$, proceeds to learn the new behavior $\tau_j = \textsc{Pick2\&Place}$.
        Subsequently, we evaluate the agent's ability of the prior behavior $\tau_{j-1}$ to determine if any forgetting has occurred.
        {\color{red} Irrelevant Action} denotes an action that results in incorrect navigation.
        `Finetuning' fails to find a target object, `Mug,' and eventually fails at the task.
        DER++ succeeds in navigating to and picking up the mug but fails to reach a microwave above the agent, also leading to task failure.
        On the contrary, our \method further succeeds in reaching the microwave, heating the mug, and putting it back on the coffee machine, leading to task success.
    }
    \label{fig:qualitative_behavior}
\end{figure*}

\subsection{The Effectiveness of Dynamically Determined Coefficients}
\label{subsec:ablation_study}

We investigate the effect of dynamically determined coefficients of our \method by fixing them with a constant value and provide the results (`\method' vs. `\method w/o D.C.' in Table~\ref{tab:sota_valid_seen}-\ref{tab:sota_valid_unseen}).
`\method w/o D.C.' assumes that the agent is \emph{always} $100\%$ confident in what it learns.
Consequently, we directly set $\gamma_a$ and $\gamma_c$ to $\alpha_a$ and $\alpha_c$ by omitting the process of dynamically determined coefficients.

We observe that the ablation of dynamically determined coefficients consistently yields performance drops in all metrics in seen and unseen environments in both \behaviorIL and \environmentIL setups, indicating the importance of the process of finding such coefficients.
This could be attributed to the fact that while logit updating with a constant coefficient helps mitigate the obsolescence of the logits to some extent, it also combines them with logits from the current model that lacks sufficient training for novel tasks, particularly during the initial phase of learning these tasks.
Consequently, this can lead to performance degradation due to incomplete knowledge of these new tasks.

\subsection{Qualitative Analysis}
\label{subsec:qualitative_analysis}
We provide qualitative examples of our \method in each \behaviorIL and \environmentIL setup by comparison with the na\"ive (\ie, Finetuning) and prior best-performing (DER++) methods.
\rev{For space's sake, we provide the qualitative example in the \environmentIL setup in Sec.~\ref{subsec:qualitative_analysis_appendix}.}

\vspace{-.75em} \paragraph{The \behaviorIL setup.}
In Figure~\ref{fig:qualitative_behavior}, the agent is evaluated for the previous behavior, $\tau_{j-1}=\rev{\textsc{Heat}}$ while learning the current behavior, $\tau_j=\rev{\textsc{Pick2\&Place}}$.
Here, the agent is required to heat a mug and put it on the coffee machine.
The agent can sequentially complete the task by 1) picking up a mug, 2) heating it using a microwave, and 3) putting it on the coffee machine.

`Finetuning' first explores the environment to find a mug.
However, it fails to recognize the mug and therefore keeps wandering in the environment, eventually leading to task failure.
Meanwhile, `DER++' succeeds in finding the mug and picking it up but forgets how to reach a microwave above the agent to heat an object.
The agent also keeps wandering in the environment and eventually, it fails at the task.
In contrast, our \method succeeds in navigating to and picking up the mug.
After grabbing the mug, our agent finds, reaches the microwave above, and successfully heats the mug.
Finally, our agent then put the heated mug on the coffee machine, as described in the instructions, which implies that our \method enables the agent to maintain the knowledge of the previous behaviors.

\section{Conclusion}
We propose two continual learning setups that learn new behaviors (\behaviorILfull, \behaviorIL) and environments (\environmentILfull, \environmentIL) continually.
Prior methods employ the storage of model logits from previous tasks but they are updated either only once or upon obtaining new logits, potentially resulting in learning with outdated data or utilizing logits from a model that has incompletely learned the new tasks.

To effectively update the logits, we propose \methodfull (\method), a simple yet effective approach that dynamically determines moving average coefficients based on the agent's confidence scores.
We observe that the \method outperforms all prior arts by noticeable margins. %, implying the efficacy of our proposed logit-update strategy.

\vspace{-.5em} \paragraph{Limitation and future work.}
While the disjoint setup operates under the assumption that tasks in streaming data are non-overlapping~\citep{parisi2019continual}, posing a stringent test for catastrophic forgetting, such non-overlapping scenarios might not always be the case in real-world scenarios.
To address this aspect, extending our proposed setups to feature overlapped tasks in streaming data, such as blurry setups~\citep{prabhu2020gdumb,bang2021rainbow} or Gaussian scheduled regimes~\citep{shanahan2021encoders,wang2022learning}, represents a promising avenue for future research.

\newpage

\section*{Ethics Statement}
\label{sec:ethics}
This work introduces continual learning setups for interactive instruction following agents and a logit-update approach to enhance the effectiveness of knowledge distillation.
While the authors do not aim for this, the increasing adoption of deep learning models in real-world contexts with streaming data could potentially raise concerns such as privacy and model robustness.
There is a possibility of these deployed deep learning models inadvertently introducing biases or discrimination, as unresolved issues like model bias persist within deep learning.
We are committed to implementing all feasible precautions to avert such consequences, as they are unequivocally contrary to our intentions.

\section*{Reproducibility Statement}
\label{sec:reproduce}
We take reproducibility in deep learning very seriously and highlight some of the contents of the manuscript that might help to reproduce our work.
We release the data splits of the proposed benchmarks in Sec.~\ref{sec:benchmark}, our implementation of the proposed method in Sec.~\ref{sec:approach}, and the baselines used in our experiments in Sec.~\ref{sec:experiment} in {\color{magenta}\url{https://github.com/snumprlab/cl-alfred}}.

\section*{Acknowledgment}
This work was partly supported by the NRF grant (No.2022R1A2C4002300, 15\%) and IITP grants (No.2020-0-01361 (10\%, Yonsei AI), No.2021-0-01343 (5\%, SNU AI), No.2022-0-00077 (10\%), No.2022-0-00113 (20\%), No.2022-0-00959 (15\%), No.2022-0-00871 (15\%), No.2021-0-02068 (5\%, AI Innov. Hub), No.2022-0-00951 (5\%)) funded by the Korea government (MSIT).

\bibliography{iclr2024_conference}
\bibliographystyle{iclr2024_conference}

\newpage
\appendix

\section*{Appendix}

\section{Extended Related Work}
\label{sec:extended_related_work}

\paragraph{Continual learning setup.} 
\label{subsec:clsetup}
We can categorize continual learning setups into Task Incremental (task-IL) and Class Incremental (class-IL) depending on whether task ID is given during inference. In task-IL, task-ID is provided during inference~\citep{aljundi2018memory, lopez2017gradient, hossain2022rethinking}, while in class-IL, task ID is not given~\citep{koh2022online, buzzega2020dark, kohonline, bang2021rainbow}. As task-ID is not provided during inference in the real world, class-IL is closer to the real setup and is more challenging because it requires classification across all classes. We focused on class-IL to deal with more realistic situations.

\vspace{-.75em} \paragraph{Continual learning methods.}
\label{subsec:clmethods}
As the need for continual learning is increasingly highlighted across various fields, researchers have proposed a wide array of continual learning methods to prevent catastrophic forgetting. (1) Replay-Based Methods~\citep{prabhu2020gdumb, koh2022online, bang2021rainbow, wu2019large, rolnick2019experience} store some stream data in episodic memory and replaying memory data during the future learning process to prevent forgetting about previous tasks. (2) Distillation-Based Methods~\citep{buzzega2020dark, kohonline, li2017learning} retain knowledge about past data by storing logits~\citep{buzzega2020dark} or models~\citep{kohonline, li2017learning} to distill knowledge. (3) Regularization-Based Method~\citep{zenke2017continual, ewc, lesort2019regularization} prevents the overwriting of important parameters by imposing a penalty on changes to these crucial parameters. 

Recently, continual learning has also been actively investigated in more challenging task setups such as video domains~\citep{zhao2021video, villa2022vclimb, park2021class}.
The focus of their work is generally on classification problems such as action recognition in a video by observing full video frames as given.
In contrast, rather than receiving predetermined frames at once, the next observations (frames) of agents are determined by the actions that the agents take, which then require the agents to plan subsequent actions to complete tasks based on the next observations.

\vspace{-.75em} \paragraph{Embodied AI.}
Embodied AI (EAI) has garnered substantial attention, and notable advancements have been made in various tasks~\citep{anderson2018vision,krantz2020navgraph,jain2019stay,habitat19iccv,deitke2020robothor,weihs2021visual,ehsani2021manipulathor,shridhar2020alfred,padmakumar2022teach,gao2022dialfred}.
For instance, visual navigation tasks necessitate that the agent use visual observation to reach designated locations~\citep{habitat19iccv} or objects~\citep{habitat19iccv,deitke2020robothor}.
Meanwhile, vision-language navigation (VLN)~\citep{anderson2018vision,jain2019stay,krantz2020navgraph} augments visual observation with natural language descriptions, enabling the agent to plan a sequence of actions based on a comprehensive understanding of multiple modalities to successfully reach the target locations.

Furthermore, the scope of EAI tasks has expanded through the inclusion of object interaction.
\citep{weihs2021visual} necessitates the agent to relocate objects to their original state by manipulating them, while \citep{ehsani2021manipulathor} requires the agent to move objects to designated locations using six degrees of freedom (6-DoF) manipulation.
\citep{shridhar2020alfred} presents natural language descriptions, which the agent must comprehend to plan a sequence of actions and utilize predictive 2D object classes to locate objects for interaction.
Meanwhile, \citep{padmakumar2022teach,gao2022dialfred} provide natural language dialogues, in which the agent must engage in reasoning to determine the appropriate course of action and complete the tasks at hand.

However, the agents evaluated in those benchmarks are typically trained using pre-existing datasets.
Given that the data collection process can be both expensive and time-consuming, it may not always be feasible to pre-collect the requisite dataset, implying the need for continual learning for the agents.

\section{Additional CL-ALFRED Benchmark Details}
\label{sec:additional_cl_alfred_benchmark_details}

\subsection{Continual Learning Setups}
\label{subsec:continual_learning_setups_appendix}
In alignment with the common practice in the prior arts~\citep{Aljundi2019OnlineCL,buzzega2020dark,shim2021online}, we assume that both setups follow an \emph{online} and \emph{disjoint} paradigm.
In an \emph{online} setup, individual samples (here, expert demonstrations) are presented sequentially rather than being available simultaneously.
While a portion of the data is retained in episodic memory, the streaming data is accessible for learning only once.
In a \emph{disjoint} setup, each task (here, each behavior and environment type) contains distinct and unrelated information from the others.
In this setup, as the agent embarks on learning new tasks, it does not receive any samples from previously encountered tasks.
Importantly, we do not rely on predefined task boundaries for training and evaluation.

\subsection{\behaviorILfull}
\label{subsec:behavior_il_appendix}
In the ALFRED benchmark, episodes comprise seven distinct behavior types: \textsc{Examine}, \textsc{Heat}, \textsc{Pick\&Place}, \textsc{Cool}, \textsc{Clean}, \textsc{Pick2\&Place}, and \textsc{Movable}.
Each behavior type presents distinct goal conditions that the agents must fulfill by learning how to achieve them.
In the \behaviorIL setup, we employ the original validation split of the ALFRED benchmark for validation and five randomly ordered sequences of behavior types for training as follows.
\begin{enumerate}[leftmargin=15pt]
    \item
    \textsc{Examine} $\xrightarrow{}$ \textsc{Heat} $\xrightarrow{}$ \textsc{Pick2\&Place} $\xrightarrow{}$ \textsc{Cool} $\xrightarrow{}$ \textsc{Pick\&Place} $\xrightarrow{}$ \textsc{Clean} $\xrightarrow{}$ \textsc{Movable}

    \item
    \textsc{Pick\&Place} $\xrightarrow{}$ \textsc{Pick2\&Place} $\xrightarrow{}$ \textsc{Clean} $\xrightarrow{}$ \textsc{Heat} $\xrightarrow{}$ \textsc{Examine} $\xrightarrow{}$ \textsc{Movable} $\xrightarrow{}$ \textsc{Cool}

    \item
    \textsc{Pick\&Place} $\xrightarrow{}$ \textsc{Examine} $\xrightarrow{}$ \textsc{Movable} $\xrightarrow{}$ \textsc{Clean} $\xrightarrow{}$ \textsc{Pick2\&Place} $\xrightarrow{}$ \textsc{Cool} $\xrightarrow{}$ \textsc{Heat}

    \item
    \textsc{Movable} $\xrightarrow{}$ \textsc{Pick2\&Place} $\xrightarrow{}$ \textsc{Examine} $\xrightarrow{}$ \textsc{Pick\&Place} $\xrightarrow{}$ \textsc{Heat} $\xrightarrow{}$ \textsc{Cool} $\xrightarrow{}$ \textsc{Clean}

    \item 
    \textsc{Clean} $\xrightarrow{}$ \textsc{Pick\&Place} $\xrightarrow{}$ \textsc{Movable} $\xrightarrow{}$ \textsc{Heat} $\xrightarrow{}$ \textsc{Cool} $\xrightarrow{}$ \textsc{Pick2\&Place} $\xrightarrow{}$ \textsc{Examine}
\end{enumerate}

\subsection{\environmentILfull}
\label{subsec:environment_il_appendix}
Episodes in the ALFRED benchmark are generated from four distinct environment types supported by AI2-THOR~\citep{ai2thor}: \textsc{Kitchens}, \textsc{Livingrooms}, \textsc{Bedrooms}, and \textsc{Bathrooms}.
Each environment type has 30 variations of the environment type that enable agents to learn behaviors in diverse room layouts and visual appearances. % and each 27, 1, and 2 environments are used for the `train,' `validation,' and `test' splits.
Similar to the \behaviorIL setup, for training, \environmentIL also employs five randomly ordered sequences of environment types as follows.
\begin{enumerate}[leftmargin=15pt]
    \item
    \textsc{Bedrooms} $\xrightarrow{}$ \textsc{Bathrooms} $\xrightarrow{}$ \textsc{Livingrooms} $\xrightarrow{}$ \textsc{Kitchens}

    \item
    \textsc{Bathrooms} $\xrightarrow{}$ \textsc{Bedrooms} $\xrightarrow{}$ \textsc{Kitchens} $\xrightarrow{}$ \textsc{Livingrooms}

    \item
    \textsc{Bedrooms} $\xrightarrow{}$ \textsc{Livingrooms} $\xrightarrow{}$ \textsc{Bathrooms} $\xrightarrow{}$ \textsc{Kitchens}

    \item
    \textsc{Bedrooms} $\xrightarrow{}$ \textsc{Bathrooms} $\xrightarrow{}$ \textsc{Kitchens} $\xrightarrow{}$ \textsc{Livingrooms}

    \item 
    \textsc{Bathrooms} $\xrightarrow{}$ \textsc{Kitchens} $\xrightarrow{}$ \textsc{Bedrooms} $\xrightarrow{}$ \textsc{Livingrooms}
\end{enumerate}

As discussed in Sec.~\ref{subsubsec:environment_il}, we observe an imbalance in the `train' and `validation' splits of the original ALFRED across the environment types: for each \textsc{Kitchens}, \textsc{Livingrooms}, \textsc{Bedrooms}, and \textsc{Bathrooms}, $11,056$, $3,456$, $3,370$, and $3,141$ episodes for the `train' split and $432$, $129$, $106$, and $153$ for the `validation' \textit{seen} split, and $468$, $146$, $120$, and $87$ for the `validation' \textit{unseen} split.

To balance them, we subsample the train and validation episodes per environment type as follows.
For the `train' split, we subsample $3,141$ episodes, leading to $12,564$ episodes in total.
For the `validation' \textit{seen} split, we subsample $106$ episodes, leading to $424$ episodes in total.
Finally, for the `validation' \textit{unseen} split, we subsample $87$ episodes, leading to $348$ episodes in total.

\section{Extended Approach}
As described in Sec.~\ref{subsec:adaptive_moving_average}, the high-level procedure of our \method is described in Algorithm~\ref{algo:sdp}.
When new samples, denoted as $x$ (representing expert demonstrations), are received, we retrieve samples, denoted by $x'$, from episodic memory.
Subsequently, we obtain the respective logits, denoted by $z$ and $z'$, from the model, denoted by $\pi_\theta$.
With these logits, we compute the gradient of the joint loss, which combines cross entropy and knowledge distillation, to update the model parameters $\theta$.

After updating $\theta$, we maintain a record of the $N$ most recent confidence scores in separate queues, denoted by $Q$, for each action and object class for $x$.
Once these recent scores are maintained, we dynamically calculate the coefficients, $\gamma_a$ and $\gamma_c$, to weight-sum the previous and current logits, denoted by $z'{old}$ and $z'$, leading to the updated logits, denoted by $z'{new}$, which are then stored in episodic memory.
For more details on the $\gamma_a$ and $\gamma_c$ calculations, kindly refer to Sec.~\ref{subsec:adaptive_moving_average}.

\renewcommand{\algorithmiccomment}[1]{\bgroup\hfill//~#1\egroup}
\begin{algorithm}[t!]
    \caption{Pseudo code for \method}
    \label{algo:sdp}
    \begin{algorithmic}[t!]
        \State \textbf{Input}
            Model $\pi_\theta$ parameterized by $\theta$,
            Memory $\mathcal{M}$,
            Training data stream $\mathcal{D}$,
            Learning rate $\mu$,
            Confidence score queues $Q$,
            Scalar $\beta$, $N$,
            Appeared action set $\mathcal{A}$,
            Appeared object class set $\mathcal{C}$
        
        \For{$(x, y) \in \mathcal{D}$}{\small\color{azure}\Comment{Samples arrive from the stream}}
            \State \textbf{Sample} $(x', y', z'_{\text{old}}) \leftarrow \text{RandomSample}(\mathcal{M})$
            {\small\color{azure}\Comment{
                Acquire triplets from the memory
            }}
            
            \State $z, z' \leftarrow f_\theta([x;x'])$
            {\small\color{azure}\Comment{
                Obtain logits from the model
            }}
            
            \State $\mathcal{L} = \text{CrossEntropyLoss}([z;z'], [y;y']) + \beta \|z' - z'_{\text{old}}\|^2_2$
            {\small\color{azure}\Comment{
                Compute the total loss
            }}
            
            \State $\theta \leftarrow \theta - \mu\cdot\nabla_\theta \mathcal{L}$
            
            \State $Q \leftarrow \text{MaintainRecentConfidences}(Q, N, z)$
            {\small\color{azure}\Comment{
                Maintain $N$ recent confidence scores
            }}
            
            \State $\gamma = \text{CalculateAdaptiveRatio}(\mathcal{A}, \mathcal{C}, Q)$
            
            \State $z'_{\text{new}} = \gamma z'_{\text{old}} + (1 - \gamma) z'$
            {\small\color{azure}\Comment{
                Update old logits by \method (Sec.~\ref{subsec:adaptive_moving_average})
            }}
            
            \State \textbf{Update} $\mathcal{M}(x',y',z'_{\text{old}}) \leftarrow \mathcal{M}(x', y', z'_{\text{new}})$
            {\small\color{azure}\Comment{
                Update logits for samples retrieved from memory
            }}
            
            \State $\mathcal{M} \leftarrow \text{ReservoirSampler}\left(\mathcal{M}, (x, y, z)\right)$
            {\small\color{azure}\Comment{
                Update Memory
            }}
        \EndFor
    \end{algorithmic}
\end{algorithm}

\section{Extended Experiment Results}

\subsection{Evaluation Metrics}
\label{subsec:evaluation_metrics}

For training and evaluation, the ALFRED dataset~\citep{shridhar2020alfred} consists of three splits; `train,' `validation,' and `test.'
Agents are trained with the `train' split and validate their approaches in the `validation' split with the ground-truth information of the tasks in those splits.
The agents are then evaluated in the `validation' and `test' split, but they do not have any access to the ground-truth information of the tasks.
As the ground-truth labels of the `test' split are not publicly available, we evaluate our agent and baselines and report the results in the validation split.

The validation split also consists of two folds: \emph{seen} and \emph{unseen} environments in which agents are/are not trained.
\emph{Seen} environments allow evaluating the task completion ability of agents in the same visual domain as training environments.
\emph{Unseen} environments further allow evaluating agents' task completion ability in \emph{different} visual domains from training environments, which is considered more challenging than \emph{seen} environments.

For the evaluation of task completion ability, the primary metric is the success rate (SR) which measures the ratio of completed tasks, indicating the task completion ability of the agent.
Another metric is the goal-condition success rate (GC) which measures the ratio of satisfied goal conditions, indicating the partial task completion ability of the agent.
We evaluate all agents' performance in terms of SR and GC in both \emph{seen} and \emph{unseen} environments and the main metric is \emph{unseen} SR.

\citep{shridhar2020alfred} also penalizes SR and GC with path-length-weighted (PLW) scores proportional to trajectory lengths.
Considering the model~\citep{kim2021agent} used in our experiments lags significantly behind human performance in terms of task completion ability (\ie, \emph{unseen} success rates), however, we focus mainly on task completion ability and leave the examination of efficiency aspects for future investigations.

\subsection{Baselines}
\label{subsec:baselines}

CLIB~\citep{koh2022online} is a method that maintains an optimal episodic memory based on the importance of each sample.
DER++~\citep{buzzega2020dark} aims to distill information about past data by storing not only images and labels but also logits, comparing them with the logits of the current model.
ER~\citep{rolnick2019experience} constructs the training batch with half of the current task stream and the other half of the data in memory, to remember past data while learning about a new task.
MIR~\citep{Aljundi2019OnlineCL} uses samples that received the most interference from previous learning, rather than randomly retrieving from memory when composing the training batch.
EWC++~\citep{ewc} prevents forgetting from parameter overwriting by penalizing changes of important parameters.
X-DER~\citep{boschini2022class} rewrites logits for the portions corresponding to the classes of past tasks to incorporate newly acquired experience information about past tasks while learning new tasks.
Following the prior methods~\citep{singh2021factorizing,pashevich2021episodic,nguyen2021look} that keep visual encoders frozen, our agent's visual encoder also remains frozen, and thus we omit the contrastive learning part~\citep{chen2020simple} in X-DER.

\subsection{Model Architecture and Training}
\label{subsec:model_architecture_and_training}

For model architecture, we adopt a recently proposed learning-based agent \citep{kim2021agent} that perceives the surrounding views and predicts a sequence of actions and object masks using factorized branches \citep{singh2021factorizing}.
Following the common practice of prior arts~\citep{shridhar2020alfred,singh2021factorizing,pashevich2021episodic,nguyen2021look,kim2021agent}, we train our agent with imitation learning, specifically behavior cloning.
We detail the architecture and training below.

\vspace{-.5em} \paragraph{Model Architecture.}
The architecture of our \method and the baselines is based on a recent approach~\citep{singh2021factorizing} that uses separate modules for effective action prediction and object localization to better address different semantic understandings from each other.
Specifically, let $y_t = f(x_t)$ be the agent that maps the input $x_t = (v_t, l, y_{a,{t-1}})$ to the output $y_t = (y_{a,t}, y_{c,t})$.
For the input, each $v_t$ and $l$ denotes the RGB images (\ie, surrounding views) and step-by-step instructions.
For the output, each $y_{a}$ and $y_{c}$ denotes the action and object class to be interacted with.

As mentioned above, the agent $f$ is comprised of two separate modules: the action prediction module $a_t = f_{action}(v_t, l, y_{a,{t-1}})$ and the object localization module $c_t = f_{class}(v_t, l, {y_{a,t-1}})$.
Both modules encode the instructions $l$ with a self-attention-based Bi-LSTM network, resulting in the attended language feature, $\hat{l}$.
To capture the correspondence between visual and textual information, we conduct point-wise convolution for $v_t$ by filters generated from $\hat{l}$, resulting in the attended visual feature $\hat{v_t}$.
The decoder of each module updates its hidden state based on the attended visual and textual features, $\hat{v_t}$ and $\hat{l}$, with the previous action $y_{a,{t-1}}$, resulting in $h^a_t$ and $h^c_t$ for the action prediction module and the object localization module.
Finally, fully connected layers in $f_{action}$ take as input $\hat{v_t}$, $\hat{l}$, $y_{a,{t-1}}$, and $h^a_t$ and predict the current action $y_{a,t}$.
Similarly, fully connected layers in $f_{class}$ take as input $h^c_t$ and predict the current object class $y_{c,t}$ with which to interact.

For more details of the modules, kindly refer to \citep{singh2021factorizing}.

\vspace{-.5em} \paragraph{Training.} % highly likely to be put in supp.
\label{subsec:training}
Following \citep{shridhar2020alfred}, we adopt imitation learning for training, specifically behavior cloning, that mimics an expert's behaviors in a teacher-forcing manner.
Formally, let $\textbf{a}$ and $\textbf{a}^*$ be a sequence of predicted actions and the corresponding ground-truth actions.
Similarly, let $\textbf{c}$ and $\textbf{c}^*$ be a sequence of predicted object classes to be interacted with and the corresponding labels.
Then each loss of the action and object class prediction is obtained by a cross-entropy loss as below:
\begin{equation}
    \begin{split}
        \mathcal{L}_{action}(\textbf{a}, \textbf{a}^*) = - \sum_{t=1}^T a_t^* \log a_t, ~~~
        \mathcal{L}_{class}(\textbf{c}, \textbf{c}^*) = - \sum_{t=1}^T \mathbbm{1}[a_t^* = \text{interaction}] \cdot c_t^* \log c_t,
    \end{split}
\end{equation}
where $T$ denotes the length of an episode that the agent conducts and $\mathbbm{1}[a_t^* = \text{interaction}]$ is an indicator function that activates when an action $a_t^*$ is an object interaction action.

In addition, \citep{ma2019selfmonitoring} adopts progress monitoring.
Formally, let $\textbf{p}$ and $\textbf{p}^*$ be a sequence of predicted progress values and the corresponding ground-truth progress values.
Then the progress loss is obtained by a mean square error (MSE) loss as below:
\begin{equation}
    \begin{split}
        \mathcal{L}_{progress}(\textbf{p}, \textbf{p}^*) = \frac{1}{T} \sum_{t=1}^T (p_t - p_t^*)^2.
    \end{split}
\end{equation}

Using them, the agent jointly minimizes the joint loss as follows:
\begin{equation}
    \label{eq:loss}
    \begin{split}
        \mathcal{L}(y, y^*) = \lambda_a \mathcal{L}_{action}(y_a, y_a^*) + \lambda_c \mathcal{L}_{class}(y_c, y_c^*) + \lambda_p \mathcal{L}_{progress} (y_p, y_p^*),
    \end{split}
\end{equation}
where $y$ indicates the output of a model including an action sequence, $y_a$, a class sequence, $y_c$, and a progress value sequence, $y_p$, for an auxiliary task.
$y_a^*$, $y_c^*$, and $y_p^*$ denote the corresponding ground-truth labels.
The loss terms are summed by the balancing coefficients $\lambda_a$, $\lambda_c$, and $\lambda_p$.

\subsection{Implementation Details}
\label{subsec:implementation_details}
For visual observation, inspired by \citep{nguyen2021look,kim2021agent,suvaansh2023multi}, we allow the agent to perceive surrounding views (in this case, 5 views from the front, left, right, up, and down directions).
For language instructions, the agents receive step-by-step instructions that describe how to accomplish the goal in detail.

For the training loss described in Section~\ref{subsec:model_architecture_and_training}, we set the balancing coefficients $\lambda_a=1.0$, $\lambda_c=1.0$, and $\lambda_p=1.0$ for our \method and the baselines.
Following~\citep{singh2021factorizing}, we augment visual observations by adopting two strategies: AutoAugment~\citep{cubuk2018autoaugment} and RGB-channel swapping.
For computational efficiency, we cache 6 types of augmented episodes per episode and choose one of them whenever we augment it.

To update the parameters of our \method and the baselines, we use the Adam optimizer with an initial learning rate of $0.001$ and a batch size of $32$ per streamed sample.
We utilize the ExponentalLR~\citep{li2019exponential} and ResetLR~\citep{loshchilov2016sgdr} schedulers with $\gamma=0.95$ and $m=10$ for our \method and the baselines except CLIB with $\gamma=0.9999$.

\subsection{Qualitative Analysis}
\label{subsec:qualitative_analysis_appendix}
We provide a qualitative example of our \method in the \environmentIL setup by comparison with the na\"ive (\ie, Finetuning) and prior best-performing (DER++) methods.

\begin{figure*}[t!]
    \centering
    \includegraphics[width=.99\linewidth]{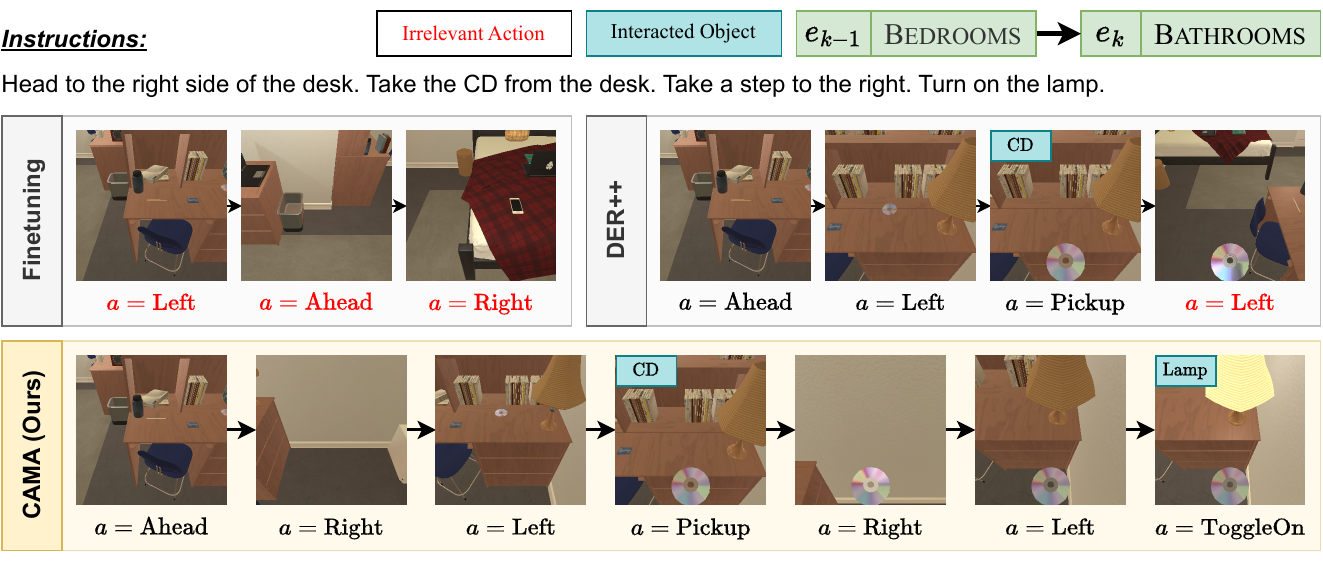}
    \vspace{-.75em}
    \caption{
        \textbf{Qualitative analysis of the proposed method (\environmentIL).}
        The agent that has already acquired knowledge of the environment $e_{k-1} = \textsc{Bedrooms}$ proceeds to learn the new environment $e_k = \textsc{Bathrooms}$.
        We then assess the agent's capability in the prior environment $e_{k-1}$ to determine whether any forgetting has occurred.
        {\color{red} Irrelevant Action} denotes an action that results in incorrect navigation.
        `Finetuning' fails to find a target object, `CD,' eventually leading to task failure.
        DER++ can navigate to and pick up the CD, but fails to turn on the lamp.
        On the contrary, our \method can also turn on the lamp and complete the task.
    }
    \vspace{-1.5em}
    \label{fig:qualitative_environment}
\end{figure*}

\vspace{-.75em} \paragraph{The \environmentIL setup.}
In Figure~\ref{fig:qualitative_environment}, the agent is evaluated for the previous environment, $e_{k-1}=\textsc{Bedrooms}$, while learning the current environment, $e_k=\textsc{Bathrooms}$.
The agent is required to examine a CD under the light of a lamp.
To complete the task, the agent needs to 1) pick up a CD and 2) turn on a lamp while holding the CD.

Similarly in the \behaviorIL setup, `Finetuning' cannot find a CD and therefore navigates to other objects irrelevant to the task, which eventually leads to task failure.
On the other hand, `DER++' successfully picks up a CD and reaches the lamp in the close vicinity.
However, the agent forgets to turn on the lamp and therefore starts to navigate to other irrelevant objects, which also leads to task failure.
In contrast, our \method can pick up the CD and navigate to the lamp as `DER++' does.
Our agent then turns on the light to examine the held CD and succeeds in the task.

\subsection{Imbalanced Scenarios in the Environment-IL Setup}
\label{subsec:imbalanced_environment_il}

\rev{To explore a data-imbalance scenario~\citep{liu2022long, he2023long}, we remove the subsampling process in Sec.~\ref{subsubsec:environment_il} for the \environmentIL setup and construct an imbalanced dataset that we name the imbalanced \environmentIL setup.
In the imbalanced \environmentIL setup, we compare our \method with the baselines and summarize the result in Table~\ref{tab:imbalanced_environment_il}.}

\rev{We observe that even with such an imbalance, \method still outperforms the baselines by noticeable margins, highlighting the efficacy of the proposed approach. In particular, we observe significant improvements of $\text{SR}_{avg}$ and $\text{GC}_{avg}$ in both valid seen and unseen splits, implying that \method achieves promising performance in both partial and full task completion.}

\begin{table*}[t!]
    \centering
    \resizebox{\textwidth}{!}{
        \begin{tabular}{@{}laaaarrrr@{}}
            \toprule
            
            \multirow{4}{*}{Model}
                 & \mcc{8}{\textbf{Imbalanced Environment-IL}} \\
                 \cmidrule{2-9}
                 & \multicolumn{4}{b}{\textbf{Valid Seen}} & \mcc{4}{\textbf{Valid Unseen}} \\
                 & \multicolumn{1}{b}{$\text{SR}_{last}\uparrow$} & \multicolumn{1}{b}{$\text{GC}_{last}\uparrow$}
                 & \multicolumn{1}{b}{$\text{SR}_{avg}\uparrow$} & \multicolumn{1}{b}{$\text{GC}_{avg}\uparrow$}
                 & \multicolumn{1}{c}{$\text{SR}_{last}\uparrow$} & \multicolumn{1}{c}{$\text{GC}_{last}\uparrow$}
                 & \multicolumn{1}{c}{$\text{SR}_{avg}\uparrow$} & \multicolumn{1}{c}{$\text{GC}_{avg}\uparrow$} \\
                             
            \cmidrule{1-9}
            
            EWC
                & $27.58 \pm 2.03$ & $38.58 \pm 2.40$ 
                & $35.44 \pm 1.42$ & $45.14 \pm 1.58$ 
                & $9.76 \pm 0.70$ & $22.94 \pm 0.61$ 
                & $12.32 \pm 1.03$ & $28.06 \pm 1.31$ 
                \\
            ER
                & $32.22 \pm 1.98$ & $43.03 \pm 1.93$ 
                & $38.87 \pm 0.29$ & $49.06 \pm 0.83$ 
                & $11.80 \pm 0.86$ & $26.43 \pm 0.84$ 
                & $13.43 \pm 0.61$ & $30.75 \pm 0.52$ 
                \\
            MIR
                & $27.58 \pm 2.03$ & $38.58 \pm 2.40$ 
                & $35.44 \pm 1.42$ & $45.14 \pm 1.58$ 
                & $9.76 \pm 0.70$ & $22.94 \pm 0.61$ 
                & $12.32 \pm 1.03$ & $28.06 \pm 1.31$ 
                \\
            CLIB
                & $24.70 \pm 2.20$ & $35.21 \pm 2.78$ 
                & $35.32 \pm 1.77$ & $44.55 \pm 1.78$ 
                & $8.24 \pm 1.14$ & $21.96 \pm 1.11$ 
                & $10.97 \pm 1.50$ & $28.03 \pm 1.11$ 
                \\
            DER++
                & $31.54 \pm 1.42$ & $42.99 \pm 1.61$ 
                & $33.01 \pm 1.97$ & $43.89 \pm 2.21$ 
                & $12.20 \pm 0.69$ & $27.28 \pm 1.23$ 
                & $11.90 \pm 1.46$ & $28.33 \pm 1.74$ 
                \\
            X-DER
                & $31.10 \pm 1.10$ & $42.83 \pm 1.13$ 
                & $32.96 \pm 1.70$ & $43.81 \pm 1.64$ 
                & $12.93 \pm 0.51$ & $27.44 \pm 0.97$ 
                & $12.49 \pm 0.95$ & $29.07 \pm 1.30$ 
                \\
            \cmidrule{1-9}  
            \method w/o D.C.
                & $\textbf{35.40} \pm \textbf{1.34}$ & $\textbf{46.55} \pm \textbf{1.40}$ 
                & $\textbf{39.54} \pm \textbf{1.30}$ & $\textbf{49.27} \pm \textbf{1.33}$ 
                & $14.19\pm 1.22$ & $\textbf{29.11} \pm \textbf{1.42}$ 
                & $16.02 \pm 1.20$ & $33.16 \pm 1.25$ 
                \\
            \textbf{\method (Ours)}
                & $32.32 \pm 1.62$ & $43.72 \pm 1.81$
                & $38.67 \pm 2.07$ & $49.09 \pm 1.90$
                & $\textbf{14.53} \pm \textbf{0.40}$ & $28.75 \pm 0.92$
                & $\textbf{17.24} \pm \textbf{1.78}$ & $\textbf{33.36} \pm \textbf{1.47}$
                \\
            \bottomrule
        \end{tabular}
    }
    \vspace{-.5em}
    \caption{
        \rev{\textbf{Comparison with state-of-the-art methods in the imbalanced \environmentIL setup.}
        The highest value per metric is in \textbf{bold}.
        We report the means and standard errors of multiple runs for random seeds.}
    }
    \vspace{-.5em}
    \label{tab:imbalanced_environment_il}
\end{table*}

\section{Discussion}
\subsection{Confidence Scores as a Good Indicator of New Logits' Quality}
\label{subsec:confidence_reason}

\rev{We use the averaged class-wise confidence score as an indicator to estimate how much new logits are informative. This is because using the confidence scores for the ground truths, which we use for logit update, allows us to estimate how well the model has learned respective classes.}

\rev{For example, If the model $p$ predicts $p(i) = 1$ for the class $i$, it implies that the model has learned the class $i$ well, \ie, it may contain ample information about the class $i$. Conversely, if the model predicts $p(i) = 0$, it implies that the model has learned the class $i$ poorly, \ie, it may contain little information about the class $i$~\citep{kohonline}.}

\rev{We can use the most \emph{single} recent confidence score as the ‘indicator,’ but such a single confidence score could be noisy during training for various reasons such as the degree of augmentation and the difficulty of a sample. To alleviate this issue, we use the $N$ recent confidence scores as an indicator of the quality of the new logits.}

\subsection{Dependency of Performance on a Task Order in Incremental Setups}
\label{subsec:task_correlation}

\rev{We agree that the performance improvements seem relatively marginal, possibly due to the large standard error of the means.
We believe this is because, in an incremental setup in embodied tasks, some tasks may share relatively many action and object classes, while others may share fewer classes.
Previously learning such shared action and object classes may help better learning the current task (\ie, forward transfer) and this implies that a model’s performance may depend on the order of the tasks (\ie, how much the model learns the shared action and object classes ahead).}

\begin{figure*}[t!]
    \centering
    \includegraphics[width=\linewidth]{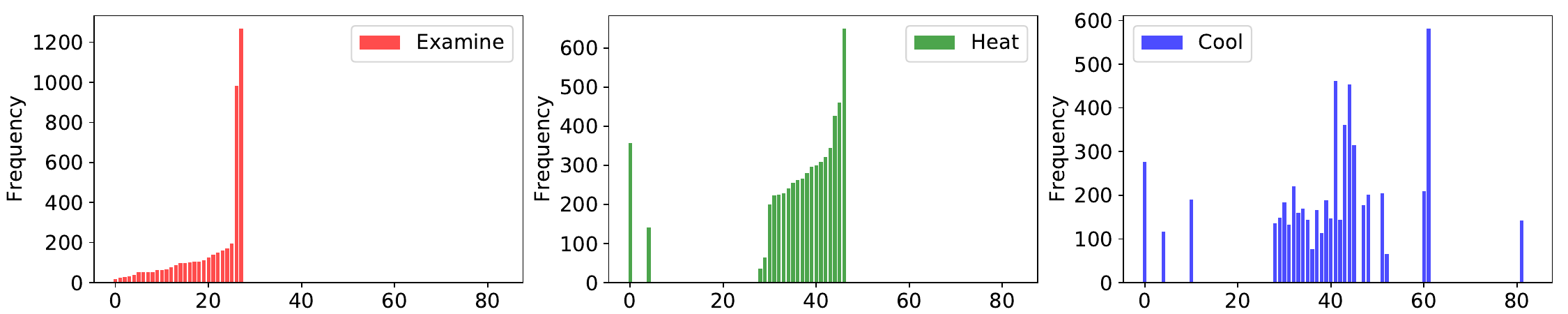}
    \vspace{-1.5em}
    \caption{
        \rev{\textbf{The frequency of objects used for each behavior in the Behavior-IL setup.} Each x-axis and y-axis denotes the index of an object and the object's frequency appearing in the corresponding behavior. While the behaviors, \textsc{Heat} and \textsc{Cool}, have several shared objects (\eg, apples, tomatoes, \etc) during task completion, the behavior, \textsc{Examine}, rarely have them with \textsc{Heat} and \textsc{Cool}.}
    }
    \vspace{-.5em}
    \label{fig:object_distrib}
\end{figure*}

\rev{For example, while learning to cool an object, learning some actions (\eg, opening/closing a fridge) and object classes (\eg, apples, tomatoes, \etc) may help next learn to heat an object as such actions and object classes can also be used for heating (\eg, heat an ‘apple,’ a ‘tomato,’ \etc by ‘opening/closing’ a microwave).
We empirically observe that for the behavior, ‘Heat,’ our agent achieves $3.70\%$ Valid Unseen SR after learning the behavior, ‘Cool,’ while it achieves zero Valid Unseen SR after learning the behavior, ‘Examine,’ which does have fewer shared object classes as illustrated in Figure~\ref{fig:object_distrib}, implying the dependency of performance on a task order.}

\end{document}